\documentclass[11pt,letterpaper]{article}
\usepackage{emnlp2016}
\usepackage{times}
\usepackage{latexsym}
\usepackage{adjustbox}
\usepackage{amsmath}%
\usepackage{amsfonts}%
\usepackage{amssymb}%
\usepackage{array}
\usepackage{graphicx} % required to import graphic files
\usepackage{color} %allows to mark some entries in the tables with color
\usepackage[caption=false,font=normalsize,labelfont=sf,textfont=sf]{subfig}
\usepackage[utf8]{inputenc}
\usepackage{booktabs}
\graphicspath{{./jpg/}}

\renewcommand{\mid}{\textit{mid} }
\newcommand{\Fig}[1]{Figure~\ref{#1}}
\newcommand{\Tab}[1]{Table~\ref{#1}}

\newcommand{\linktodatasets}{http://dx.doi.org/10.17632/cdcztymf4k.1}

\usepackage{ragged2e}
\newcolumntype{P}[1]{>{\RaggedRight\hspace{0pt}}p{#1}}

\emnlpfinalcopy % Uncomment this line for the final submission
\title{Automatically Annotated Turkish Corpus for Named Entity Recognition and Text Categorization using Large-Scale Gazetteers}
\author{H. Bahadir Sahin, {\bf Caglar Tirkaz}, {\bf Eray Yildiz}, \\ {\bf Mustafa Tolga Eren},  {\bf Ozan Sonmez} \\
	Huawei Turkey Research and Development Center, Umraniye, Istanbul, Turkey \\
	{eray.yildiz@huawei.com} \\ 
	{{hbahadirsahin, caglartirkaz, tolgaeren, osonmez}@gmail.com}
}

\begin{document}	
\maketitle	
\begin{abstract}

Turkish Wikipedia Named-Entity Recognition and Text Categorization (TWNERTC) dataset is a collection of automatically categorized and annotated sentences obtained from Wikipedia. We constructed large-scale gazetteers by using a graph crawler algorithm to extract relevant entity and domain information from a semantic knowledge base, Freebase\footnote{https://www.freebase.com/}. The constructed gazetteers contains approximately 300K entities with thousands of fine-grained entity types under 77 different domains. Since automated processes are prone to ambiguity, we also introduce two new content specific noise reduction methodologies. Moreover, we map fine-grained entity types to the equivalent four coarse-grained types, \textit{person, loc, org, misc}. Eventually, we construct six different dataset versions and evaluate the quality of annotations by comparing ground truths from human annotators. We make these datasets publicly available to support studies on Turkish named-entity recognition (NER) and text categorization (TC).

% (HTC and HEC) which are based on MIT license. 
	
\end{abstract}

\section{Introduction}
	
Named-entity recognition (NER) is an information extraction (IE) task that aims to detect and categorize entities to pre-defined types in a text. 
On the other hand, the goal of text categorization (TC) is to assign correct categories to texts based on their content. 
Most NER and TC studies focus on English, hence accessing available English datasets is not a issue. 
However, the annotated datasets for Turkish NER and TC are scarce. 
It is hard to manually construct datasets for these tasks due to excessive human effort, time and budget.
In this paper, our motivation is to construct an automatically annotated dataset that would be very useful for NER and TC researches in Turkish.

The emergence of structured and linked semantic knowledge bases (KBs) provide an important opportunity to overcome these problems. Approaches that leverage such KBs can be found in literature \cite{heck2013leveraging,gerber2013real,hoffart2011robust,mendes2011dbpedia}. 
However, using the structured data from KBs is a challenging task for linking named entities and domains to raw texts due to ambiguous texts and named entities  \cite{cucerzan2007large}. 

In this work, we publish TWNERTC dataset in which named entities and categories of sentences have been automatically annotated. We use Turkish Wikipedia dumps as the text source\footnote{https://dumps.wikimedia.org/} and Freebase to construct a large-scale gazetteers to map fine-grained types to entities. To overcome noisy and ambiguous data, we leverage domain information which is given by Freebase and develop domain-independent and domain-dependent methodologies.
All versions of datasets can be downloaded from our project web-page\footnote{\linktodatasets}. 
Our main contributions are (1) the publication of Turkish corpus for coarse-grained and fine-grained NER, and TC research, (2) six different versions of corpus according to noise reduction methodology and entity types, (3) an analysis of the corpus and (4) benchmark comparisons for NER and TC tasks against human annotators. To the best of our knowledge, these datasets are the largest datasets available for Turkish NER ad TC tasks. 

The rest of the paper is organized as follows: In Section 2, we briefly investigate the literature about NER, TC and datasets which are used in these research. In Section 3, we explain the construction of large-scale gazetteers by using Freebase. In Section 4, we explain how to use the gazetteers to automatically annotate and categorize Wikipedia texts to construct datasets along with dataset statistics, and noise reduction methodologies. Our evaluation about the quality of these constructed datasets are reported in Section 5.

\section{Related Work}

% EMNLP için literatür

Named Entity Recognition (NER) and Text Classification (TC) are well-researched NLP tasks relevant to large amount of information retrieval and semantic applications. TC research predates to '60s; however, it is accepted as a research field in '90s with the advances in technology and learning algorithms \cite{sebastiani2002machine}. On the contrary, the classical NER task is defined in MUC \cite{chinchor1998overview} and CoNLL \cite{tjong2003introduction} conferences with coarse-grained entity types: person, location, organization and misc. In addition, few studies address the problem of fine-grained NER where the challenge is to capturing more than four entity types \cite{pasca2006organizing,ekbal2010assessing,yogatama2015embedding}. 

As research on NER has been pushing the limits of automated systems performing named-entity recognition, the need for annotated datasets and benchmarks is also increasing. Knowledge bases are important for NLP researches, since they provide a structured schema of topics that can be used to annotate entities with fine-grained types and/or categorize raw texts into related domains. 
	
Steinmetz et al. \cite{steinmetz2013statistical} published benchmark evaluations that compare three datasets that use semantic information from KBs: DBpedia Spotlight \cite{mendes2011dbpedia}, KORE50 \cite{hoffart2011robust,yosef2011aida} and the Wikilinks Corpus \cite{singh2011large}. These datasets are in English and constructed with the aim of evaluating the performance of NER systems. The authors present the statistics of each dataset and baseline performances of various algorithms. There are other methodologies which leverages KBs to named entity extraction and linking; however, most of them are not available to  public \cite{hoffart2011robust,heck2013leveraging}.

Constructing a comprehensive dataset for TC is tougher than NER since there is no limit for the number of categories that are represented in such sets. In general, there are many TC datasets available in English for many different problems such as sentiment analysis \cite{liu2015automated} and categorizing gender \cite{mukherjee2010improving}. The largest and the most popular dataset among them is Reuters Corpus Volume 1 (RCV1) which consists of manually categorized 800K news stories with over 100 sub-categories under three different main categories \cite{rose2002reuters}. This version the dataset has problems with document categories and suffers from lack of documentation about the dataset. Lewis et al. propose an improved version of this dataset with reduced categorization mistakes and provide a better documentation \cite{lewis2004rcv1}. 

Research on Turkish NER and TC are very limited compared to English and several other languages.  The main reason is the lack of accessibility and usability of both Turkish NER and TC datasets. The most popular Turkish NER dataset is introduced by Gökhan et al. \cite{tur2003statistical}. This dataset contains articles from newspapers, approximately 500K words, and is manually annotated with coarse-grained entity types. Tatar and Çiçekli propose another coarse-grained NER dataset \cite{tatar2011automatic}; however, it contains only 55K words which makes this dataset to less preferable than previous dataset. More recent studies focus on Turkish NER in social media texts \cite{onal2015named,kuccuk2014named,demir2014improving,celikkaya2013named}. Due to the research focus in the field, several Twitter-based coarse-grained NER datasets are published \cite{kuccuk2014named,kuccuk2014experiments,tantugrecognizing}. According to our knowledge, there is no literature available regarding to fine-grained NER in Turkish.

Turkish TC researchers tend to construct their own, case specific datasets in general \cite{amasyali2006automatic}. The newspapers are the main text source for such studies since they are easy to obtain and classify manually \cite{akkus2013categorization,toraman2011developing,kilincc2015ttc}. 
When the amount of annotated data is considered to train state-of-the-art learning algorithms, aforementioned Turkish datasets suffer from the lack of enough data. The main bottlenecks are  requires human effort and time constraint, which limits the size and scope of the constructed datasets. In contrast, our aim is to provide larger, more comprehensive and insightful Turkish datasets for both NER and TC by using knowledge bases to create large-scale gazetteers and eliminating human factor in the annotation process.

In the next section, we will explain our dataset construction methodology starting with building gazetteers by using Freebase and Wikipedia. We will investigate how to build a graph crawler algorithm to crawl the knowledge in Freebase and to map its entity types and domain information into the raw texts automatically. We will also discuss noise reduction methods and propose three different versions of the dataset.

\section{Dataset Construction Methodology}
 
\subsection{Constructing Gazetteers}

Gazetteers, or entity dictionaries, are important sources for information extraction since they store large numbers of entities and cover vast amount of different domains. 
KBs use a graph structure that is used to represent thousands of films and/or millions of songs in gazetteers. Hence, such graph structures can be efficient to construct large-scale gazetteers.

In our work, we use Freebase as the KB since it provides both quality and quantity in terms of entity types and domains for Turkish.
Freebase has 77 domains that cover many different areas from music to meteorology with approximately 50M total entities. 
Among them, Turkish has 300K entities, and approximately 110K of them have a link to corresponding Turkish Wikipedia page. 

By using KBs, one can eliminate the necessity of creating semantic schema design, collecting data and manually annotating raw texts. However, such large entity lists contain ambiguous and inaccurate information that can impact the quality. For instance, both film and boat domains contain ``Titanic" in their entity lists, and such ambiguity could create false links. Considering the number of entities, the importance of disambiguating named entities becomes more important.

Our work is inspired from Heck et al.\cite{heck2013leveraging} who employ a method that takes advantage of user click logs of a web search engine in order to improve precision of gazetteers while maintaining recall. 
% Their approach compares the click distribution of an entity to aggregate click distributions of random phrases to obtain a quality score for each entity. Then, they eliminate all entities from the gazetteers that have low confidence scores.
Since we do not have any sources to access such user click logs, we depend on attributes, such as entity domains, types and properties, that Freebase provides in order to improve the quality of our gazetteers. Note that a named entity's domain, type and property are represented as \textit{/domain/type/property} in Freebase.

Freebase distinguishes entities that exist in more than one domain by unique \textit{machine id} (\mid). 
For instance, Titanic is an entity in both film and boat domains with two \textit{mids} depending on the domain. 
On the other hand, there are cases in which same \textit{mids} can be associated with multiple domains. For instance the \mid of Titanic in the film domain is also used in the award domain. These cases occur when domains are closely related.
Note that \mid is language independent. Hence, an entity has the same \mid regardless of its language; however, the information related to that entity can differ, e.g., missing equivalence of entities, translation differences.

Domains covered in Freebase have large amount of entities and related descriptive texts. However, for Turkish, we need to filter or merge some domains due to insufficient number of related raw texts. For instance, we merge ``American football", ``cricket", ``ice hockey" domains under already existing \textit{sports} domain. 

\begin{figure*}[!ht]
	\centering
		\includegraphics[width=0.9\textwidth]{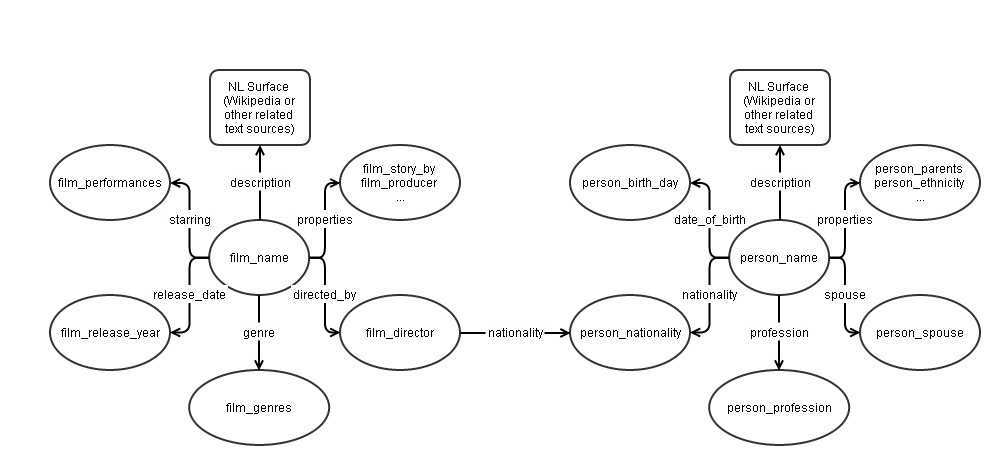}
	\caption{Examples of entity types and relations between them.}
	\label{fig:Freebase_Graph_Structure}
\end{figure*}

During annotation, due to ambiguous cases in Freebase, we first compute the domain and the entity type distribution of directly related (first order) relations of the selected named entity. 
Among the candidates, the entity type that contains the most first order relations in the knowledge graph is selected as the type of the entity. 
This ensures that the most informative entity type is selected for the annotated entity. 
For instance, while annotating the Wikipedia page of ``Titanic (film)" the possible candidates are \textit{/film/film} and \textit{/award/award\_winning\_work}. Our method chooses \textit{/film/film} as the domain and type of the entity since it has more information compared to its competitor.

After determining the entity type of each entity, we form large-scale gazetteers for Turkish which contain 300K named-entities. Each entity in the gazetteer has its Wikipedia description, determined type and 1st-order relations. Note that some entities may not have Wikipedia description due to several reasons, e.g. song names or deleted Wikipedia pages; however, we use such entities in the annotating process.

\section{Automatic Annotation Generation}

\subsection{Fine-Grained Annotations}

A knowledge graph consists of nodes and edges between nodes which are defined by the schema of a domain. 
Nodes are entities and edges represent relations between nodes, which are \textit{properties} in Freebase. Examples of entities and relations are shown in \Fig{fig:Freebase_Graph_Structure}, for ``film" and ``people" domains. The central node having the most number of relations in the film domain is the ``film\_name". We call directly connected relations to the central node as first-order relations. 
Name of the director who directed the movie is ``film\_director" relation. Since directors are people, they have also relations in ``person" domain. 
Through such relationship we can get second-order relations about a movie. 
We benefit this graph structure along with the raw texts to create multi-purpose, annotated and categorized corpora. Inspired by the approach of \cite{heck2013leveraging}, we create a graph crawling algorithm that is capable of categorizing sentences into Freebase domains and annotate named entities within the sentences with Freebase types and properties. 
This set of annotations are the \textit{Fine-Grained Annotations} (FGA). 
Our method consists 5 steps and is applicable to both English and Turkish (with slight changes for sentence processing). 

\begin{table*}[ht!]
	\centering
	\begin{tabular}{lcccc}
		\toprule
		 & TWNERTC-DI & TWNERTC-DD \\ 
		\midrule
		\# Of Sentences  & 695949 & 696785 \\ 
		Largest Domains (\# sentences)   & People(137789) & People(137907)\\ 
		Smallest Domains (\# sentences)  & Exhibitions(132) & Exhibitions (132)\\ 
		\# of Tokens (with punctuation)  & 13445651 & 13458930 \\ 
		\# of Tokens (without punctuation)  & 10949015 & 10960244 \\ 
		\# of Tagged Tokens  & 2314333 & 2255320 \\ 
		\# of Unique Entity Types  & 3150 & 3447 \\ 
		Largest Domains (\# unique entities)  & People(1536) & People (1011)\\ 
		Smallest Domains (\# unique entities)  & Physics(58) & Physics(12)\\ 
		\bottomrule
	\end{tabular}
    \caption{Statistics of TWNERTC datasets after Domain Independent (DI) and Domain Dependent (DD) post-processing.}
	\label{tab:HEC_HTC_Statistics_Noise_Reduction}
\end{table*}

\begin{enumerate}
% 	\item \textbf{Initialization}\\
	\item Select the central node from gazetteers. This node is ``Central Pivot Node" (CPN). It is the main entity-type of the domain, e.g. film names in film domain.
% 	\item \textbf{Retrieve description of the entity}\\
	\item Retrieve the descriptions provided by Wikipedia or Freebase. 
	If Wikipedia has the corresponding page, fetch full texts from dump. Else if only Freebase description is available, use the description. Otherwise, return to the first step. Since several Turkish Wikipedia texts are direct copy of English version, language detection is applied on all texts and English texts are eliminated.
% 	\item \textbf{Annotate first order relations}\\
	\item Extract sentences from raw texts and annotate sentences by longest-string pattern matching where the first-order relations of CPN are used resulting in IOB style annotation. We use Aktaş and Çebi's sentence detection method to extract sentences from full texts \cite{aktacs2013rule}.
% 	\item \textbf{Annotate second order relations} \\
	\item Extend annotation process with second-order relations.
% 	\item \textbf{Assign a domain label}
	\item Categorize sentences by selecting the domain of entities that is used the most. 
\end{enumerate}

One should consider that higher order relations can create a long chain of relationships. 
In our work, we limit this chain with the second-order relations, since further relations provide less information while increasing ambiguity and inconsistency of automated annotations. 
	
\subsection{Statistics of the FGA}
%We processed approximately 300K entities for Turkish that we have crawled from Freebase. Among these entities, 110K entities have Turkish description (from Freebase or Wikipedia dump) which makes them CPN candidates. The remaining entities are used for annotations if any CPN has them any relationships with them. 

TWNERTC contains 300K entities in total of which 110K have Turkish descriptions (from Freebase or Wikipedia dump). 
There are totally 700171 annotated sentences from 49 different domains with the largest and smallest domains being \textit{people} (139K sentences) and \textit{fashion} (493 sentences) subsequently. 
% HTC contains 13503399 and 10997037 tokens with and without punctuations respectively. 
TWNERTC consists of 10997037 tokens without punctuation and among these tokens approximately 2M of them are annotated. 16K of the tags are unique which results in approximately 332 unique entity type per domain on the average. 
Note that, even if a sentence is categorized into the \textit{film} domain, it can contain entities from other domains, such as \textit{person} and \textit{time}. \textit{Location} is the domain having the highest number of unique tags (1051) whereas \textit{physics} has the least amount of unique tags (15).

%We obtain 700171 sentences with 49 different domains and corresponding annotations that have at least one annotated entity in total. While the biggest domain is \textit{people} with 139K sentences, the smallest domain is fashion with 493 sentences. The rest of sentence domains differs between 1K to 100K. HTC contains 13503399 word tokens with punctuations, and 10997037 without punctuations. Among these tokens approximately 2M of them are tagged and 16K of the tags are unique which makes approximately 332 unique tags per domain on average. Note that even if a sentence is categorized in \textit{film} domain, it can contain entities from other domains, such as \textit{person}, \textit{time}. Sentences with \textit{location} domain has the most unique tags with 1051 and \textit{physics} domain has only 15 unique tags. 

\subsection{Disambiguate Noisy Entity Types}
We refine the gazetteers to minimize the effect of noise in the generated annotations and assigned domains. However, due to the nature of automated algorithm, we find that entity annotations have still inconsistent or missing types while categorization process is resulted with more accurate in general. 
%Hence, we decide to reduce noisy entities, leaving sentence categories as is. 
In order to reduce remaining noise, we apply both \textit{domain dependent} and \textit{domain independent} noise reduction. 
% Both of the approaches modify existing annotations by changing the annotations associated with an entity type with the most common annotation used to annotate that entity type. 
% The difference between the two approaches is that the domain dependent approach performs preprocessing within each domain whereas domain independent approach does not. 
Domain dependent approach finds the most common entity type of every entity according to the domain of the sentences. Then, entities are re-annotated with the common entity types. Domain independent approach follows the same process without using domain information while eliminating the noisy information. 
Statistics about these two versions of the dataset are presented in \Tab{tab:HEC_HTC_Statistics_Noise_Reduction}.

\subsection{Transform FGA to Coarse-Grained Annotation}

\begin{table*}[!ht]
	\centering
	\begin{tabular}{lcccc}
		\toprule
		& People  & Location & Organization & Misc   \\ 
		\midrule
		TWNERTC & 353385  & 296318   & 75940        & 973477 \\ 
		TWNERTC-DI  & 345545  & 331943   & 90673        & 1172214 \\ 
		TWNERTC-DD & 394291  & 361728   & 99075        & 1046818 \\ 
		\bottomrule
	\end{tabular}
    \caption{Number of tokens for each entity type in TWNERTC with CGA transformation}
	\label{tab:htc_hec_statistics_cga}
\end{table*}

FGA provides fine-grained annotations with many detailed entity types and properties. However, the amount of different entity types affects the learning algorithms negatively. 
Moreover, it is hard to evaluate the quality of the annotations when most works in literature performs coarse-grained entity recognition. Thus, we provide a coarse-grained version of the FGA datasets.
In order to transform FGA to coarse-grained annotation (CGA), we map each fine-grained entity type in each domain to a coarse-grained entity type, i.e. \textit{person}, \textit{organization}, \textit{location} and \textit{misc}. 
We keep the IOB notations in types while converting the type.
In this process, we eliminate several domains, such as meteorology, interests and chemistry, due to the lack of types that can be mapped to a coarse-grained version. This elimination process leaves 25 unique domains in CGA-based datasets.

In coarse version of the dataset, there are approximately 380K sentences. Similar to FGA statistics, the \textit{people} domain has the highest number of sentences (104508) while \textit{law} domain has the least number of sentences (454) among all 25 domains. The number of tokens is 7326286 with punctuation. 
Among these tokens, 851123 of them are annotated. Unlike FGA, CGA has only 4 types for each domain. 19 of the domains contains all 4 types whereas the remaining domains might contain 2 or 3 entity types such as geography and food. 

We also transform the post-processed datasets we introduced in previous chapter to CGA. \Tab{tab:htc_hec_statistics_cga} presents the details of created corpora with CGA. In total, we publish six datasets (one original and two post-processed with FGA, one original and two post-processed with CGA)\footnote{\linktodatasets}.  
	
\section{Evaluation}

To experimentally evaluate TWNERTC, five human annotators categorize and annotate test sets that are sampled from the datasets. We create six test sets (3 CGA, 3 FGA) with 10K-word for NER and one test with 2K sentences for TC. We compare annotations and domains of these sets with manually created ground truths.

For NER evaluation, we follow two different approaches for coarse-grained and fine-grained versions. While we evaluate the CGA versions against manually annotated ground-truths, it is an almost impossible to evaluate FGA versions. Hence, we train a fine-grained NER model to predict top-5 possible entity types for all entities and human annotators create ground-truths by using these predictions. For both cases, we extract 10K-word test sets for all three versions (original and post-processed). Note that, test sets are not identical but randomly selected sentences from the datasets.  In addition, we exclude IOB tags since we prioritize evaluating entity type agreement in our evaluation results. 

For evaluating CGA versions, given the sentence and the corresponding automatically created annotation, annotators are allowed to change any entity type with one of the five possible types, i.e. person, organization, location, misc or O (out). Finally, we merge the results of all annotators such that if at least 3 annotators agree on the same type for an entity, that type is the ground-truth. If there is no agreement, we keep the entity type as it is.

For evaluating FGA, we use a fine-grained entity recognizer, FIGER \cite{ling2012fine}\footnote{https://github.com/xiaoling/figer}, to create ground-truths. It is not an ideal evaluation approach since FIGER is designed for English and have not been tested for other languages according to our knowledge. However, more than thousands of different entity types are available in TWNERTC, and it is impracticable to ask human annotators to construct such fine-grained ground-truths manually from scratch. Therefore, we train a Turkish fine-grained model by using all remaining sentences and predicted possible types for entities in the test sets. Then, we ask human annotators to rank types of an entity from the most relevant to the least relevant. They are also allowed to suggest alternative types different than the given choices. 

\begin{table*}[!ht]
    \centering
	\begin{tabular}{lcccc}
		\toprule
        & TWNERTC & TWNERTC-DI &TWNERTC-DD \\
		\midrule
		\# of Annotated Entities & 2376 & 1858 & 1965 \\ 
		\# of Ground-Truths for Entities   & 2891  & 2653 & 2 771 \\ 
        \# of Added Entity Type   & 872 & 958 & 926 \\ 
        \# of Removed Entity Type & 537 & 163 & 120 \\ 
        \# of Changed Type  & 564 & 278  & 198 \\ 
        \# of Same Type	& 1275 & 1417 & 1647 \\
		\bottomrule
	\end{tabular}
    \caption{Automated vs. Manual comparison for Coarse-Grained TWNERTC.}
    \label{tab:htc_hec_evaluation_ner}
\end{table*}

We randomly sample 2K sentences from the unmodified corpus as TC test set. We train an internal classification algorithm with the rest of sentences and get top five predicted domains for the test sentences. We present predicted categories to human annotators and ask them to rank domains from the most relevant to the less relevant. They are also allowed to suggest different domains among the 49 domains which are represented in full corpus. Finally, we rank domains of each test sentence according to the annotator agreement and form the test set such that each sentence has 5 possible domains where first domain is the most relevant domain.

\subsection{Evaluation Results for Coarse-Grained Annotations}

In \Tab{tab:htc_hec_evaluation_ner}, we present the number of entity types exist in automatically and manually annotated sets with the number of changes that annotators have made. We define changes from type \textit{O} to any other type is an addition and opposite of this action is a removal. \textit{Misc} is the most added, removed and changed type by annotators. It is an acceptable outcome since this specific type covers vast amount of entities except \textit{person}, \textit{location} and \textit{organization}. We do not consider the number of added entity types as a major problem, since our gazetteers do not have infinite information and we have future plans to improve it. However, miss-annotated types are the real danger since if the amount of such mistakes increase, performance of the learning algorithms is affected negatively. In "change" type of , annotators change \textit{misc} type to mainly \textit{organization} and \textit{person}. 

\begin{table}[!ht]
    \resizebox{\columnwidth}{!}{\begin{minipage}{\columnwidth}
    \centering
	\begin{tabular}{lcccc}
		\toprule
        & Precision & Recall & F1-Score \\
		\midrule
		Person & 0.90 & 0.86 & 0.88  \\ 
        Person (DI) & 0.98 & 0.96 & 0.97  \\ 
        Person (DD) & 0.99 & 0.98 & 0.98  \\ 
		Org.  & 0.84 & 0.76 & 0.79   \\ 
        Org. (DI) & 0.94 & 0.85 & 0.9   \\ 
        Org. (DD) & 0.91 & 0.86 & 0.88   \\ 
        Loc. & 0.91 &  0.80 & 0.85   \\
        Loc. (DI) & 0.93 &  0.82 & 0.87   \\
        Loc. (DD) & 0.96 &  0.92 & 0.93   \\
        Misc & 0.79 &  0.59 & 0.67   \\ 
        Misc (DI) & 0.84 &  0.67 & 0.74   \\ 
        Misc (DD) & 0.87 &  0.71 & 0.78   \\ 
        Average & 0.81 &  0.72 & 0.76   \\ 
        Average (DI) & 0.89 &  0.76 & 0.82    \\ 
        Average (DD) & 0.91 &  0.79 & 0.84   \\ 
		\bottomrule
	\end{tabular}
    \caption{Evaluation for Coarse-Grained TWNERTC, TWNERTC-DI and TWNERTC-DD}
    \label{tab:htc_evaluation_scores_ner}
	\end{minipage}}
\end{table}

As a conclusion, among the automatically annotated coarse-grained entity types, there are \%76 matching ratio without \textit{O} tags and manually added types. Additionally, we present precision, recall and F-score values in \Tab{tab:htc_evaluation_scores_ner}. The dataset with domain-dependent post-process provides better NER performance compared to other two versions in general. Moreover, both post-processing methods improve the performance compared to the original dataset. Further, it can be observed that \textit{misc} type is has the lowest F-Score among all types in all versions; however, this result is expected since \textit{misc}'s coverage is larger than the other three types and makes it more vulnerable to mismatches. 

\subsection{Evaluation Results for Fine-Grained Annotations}

As we discuss earlier, we train an external fine-grained algorithm, FIGER, to create Turkish NER models for all versions of the TWNERTC. By using the resulted models, we take five possible predicted type for each entity that are represented in test sets, and provide them as ground-truths to five human annotators. Obviously, FIGER designed to solve fine-grained NER in English; however, to evaluate the automated fine-grained types in a reasonable time, we leverage predictions of this algorithm.

\begin{table}[!ht]
    \resizebox{\columnwidth}{!}{\begin{minipage}{\columnwidth}
    \centering
	\begin{tabular}{lcccc}
		\toprule
         & Strict & L. Macro & L.Micro \\
		\midrule
		TWNERTC  &   0.274 &  0.393  & 0.366   \\ 
        TWNERTC-DI  &  0.344 & 0.494  & 0.476   \\ 
        TWNERTC-DD  &  0.332 & 0.472 & 0.463  \\ 
		\bottomrule
	\end{tabular}
    \caption{Strict, loose macro, and loose micro F1-scores on test sets}    
    \label{tab:htc_hec_evaluation_fine_grained_training}
	\end{minipage}}
\end{table}

In \Tab{tab:htc_hec_evaluation_fine_grained_training}, we present the F1-scores of trained models on each test set. We provide these scores to give a better insight to researchers about ground-truths. Note that, strict represents the original F1-score formula, while loose macro and loose micro scores represent variations of the same formula \cite{ling2012fine}. It can be observed that the model trained with original TWNERTC performs poorly compared to post-processed versions. Since domain independent (DI) noise reduction method ensures that an entity can have only one type, it improves the performance more than the domain dependent (DD) method. 

\begin{table}[!ht]
    \resizebox{\columnwidth}{!}{\begin{minipage}{\columnwidth}
    \centering
	\begin{tabular}{lcccc}
		\toprule
         & Top-1 & Top-3 & Top-5 \\
		\midrule
		TWNERTC  &  \%22.1 & \%40.3   & \%55.3  \\ 
        TWNERTC-DI  &  \%29.2 & \%49.7 & \%59.5   \\ 
        TWNERTC-DD  &  \%27.8 & \%52.5  & \%60.9   \\ 
		\bottomrule
	\end{tabular}
    \caption{Evaluation for Fine-Grained TWNERTC}    
    \label{tab:htc_hec_evaluation_fine_grained_eval}
	\end{minipage}}
\end{table}

\Tab{tab:htc_hec_evaluation_fine_grained_eval} presents the human annotators evaluation on automated fine-grained NER datasets, given FIGER predictions as possible ground-truths. Annotators rank the provided ground-truths and we check their ranking agreements. Eventually, top-1 agreement is hard to fulfill since our gazetteers contains thousands of entity types, and an entity may have more than ten different possible options. On the other hand, top-5 agreements provide promising results considering the amount of possible ground-truths.  

\subsection{Evaluation Results for TC}

Original TWNERTC contains 49 different domains. The number of domains causes ambiguities in categorization among sentences depending on the context understanding. Thus, we evaluate three levels of accuracy. From \Tab{tab:htc_hec_evaluation_tc}, we can see that automatically assigned domains have relatively low direct matches with annotators according to top-1 score. However, when we observe top-3, accuracy scores are doubled. Furthermore, error rates are less than \%2 when all five ground truths are considered. Note that in top-3 and top-5, we only considered whether automatically assigned domain exists or not in the ground truth list.

\begin{table}[!ht]
    \resizebox{\columnwidth}{!}{\begin{minipage}{\columnwidth}
    \centering
	\begin{tabular}{lcccc}
		\toprule
         & Top-1 & Top-3 & Top-5 \\
		\midrule
		TWNERTC  &  \%46.4 & \%90.0  & \%97.5   \\ 
		\bottomrule
	\end{tabular}
    \caption{Text categorization evaluation for TWNERTC}    
    \label{tab:htc_hec_evaluation_tc}
	\end{minipage}}
\end{table}

Amount of the difference between from top-1 to top-3 scores is mainly caused by annotators' different understanding of the sentence context. For instance, a sentence about Lionel Messi's birth location is categorized as \textit{people} in the test set. Whereas ground-truths start with \textit{soccer} and followed by \textit{people} and \textit{location}. In addition, similar domains, such as \textit{sports} and \textit{soccer}, are also the reason of this difference. 

In overall, results validates that automatically assigned domains are likely similar to what human annotators suggest to corresponding sentences. However, while automation process is limited to the context, humans can use their knowledge when suggesting domains. Hence, low accuracy in top-1 score is not a mistake of the methodology but a shortcoming in the process. 

\section{Conclusion and Future Research Directions}

We have described six, publicly available corpora for NER and TC tasks in Turkish. The data consists of Wikipedia texts which are annotated and categorized according to the entity and domain information extracted from Freebase. We explain the process to construct the datasets and introduce methodologies to eliminate noisy and incorrect data. We provide comprehensive statistics about dataset content. We analyzed subsets from these datasets and evaluate automatically created annotations and domains against manually created ground-truths. The final results show that automatic annotations and domains are quite similar to the ground-truths.

The obvious next step is to develop learning algorithms for NER and TC tasks to find baselines using traditional machine learning algorithms, and extending these baselines with approaches. Since TWNERTC provides a vast amount of structured data for researchers, deep learning methods can be exploited to solve fine-grained NER problem.  

\section{Acknowledgments} 
This project is partially funded by 3140951 numbered TUBITAK-TEYDEB (The Scientific and Technological Research Council of Turkey – Technology and Innovation Funding Programs Directorate).	
	
{\small
	\bibliographystyle{emnlp2016}
	\bibliography{./ref}

\begin{thebibliography}{}

\bibitem[\protect\citename{Akkuş and Çakıcı}2013]{akkus2013categorization}
Burak~Kerim Akkuş and Ruket Çakıcı.
\newblock 2013.
\newblock Categorization of turkish news documents with morphological analysis.
\newblock In {\em ACL (Student Research Workshop)}, pages 1--8.

\bibitem[\protect\citename{Akta{\c{s}} and {\c{C}}ebi}2013]{aktacs2013rule}
{\"O}zlem Akta{\c{s}} and Yal{\c{c}}{\i}n {\c{C}}ebi.
\newblock 2013.
\newblock Rule-based sentence detection method (rbsdm) for turkish.
\newblock {\em International Journal of Language and Linguistics}, 1(1):1--6.

\bibitem[\protect\citename{Amasyal{\i} and Diri}2006]{amasyali2006automatic}
M~Fatih Amasyal{\i} and Banu Diri.
\newblock 2006.
\newblock Automatic turkish text categorization in terms of author, genre and
  gender.
\newblock In {\em Natural Language Processing and Information Systems}, pages
  221--226. Springer.

\bibitem[\protect\citename{Chinchor}1998]{chinchor1998overview}
Nancy~A Chinchor.
\newblock 1998.
\newblock Overview of muc-7/met-2.

\bibitem[\protect\citename{Cucerzan}2007]{cucerzan2007large}
Silviu Cucerzan.
\newblock 2007.
\newblock Large-scale named entity disambiguation based on wikipedia data.
\newblock In {\em EMNLP-CoNLL}, volume~7, pages 708--716.

\bibitem[\protect\citename{Demir and Özgur}2014]{demir2014improving}
Hakan Demir and Arzucan Özgur.
\newblock 2014.
\newblock Improving named entity recognition for morphologically rich languages
  using word embeddings.
\newblock In {\em Machine Learning and Applications (ICMLA), 2014 13th
  International Conference on}, pages 117--122. IEEE.

\bibitem[\protect\citename{Ekbal \bgroup et al.\egroup
  }2010]{ekbal2010assessing}
Asif Ekbal, Eva Sourjikova, Anette Frank, and Simone~Paolo Ponzetto.
\newblock 2010.
\newblock Assessing the challenge of fine-grained named entity recognition and
  classification.
\newblock In {\em proceedings of the 2010 Named Entities Workshop}, pages
  93--101. Association for Computational Linguistics.

\bibitem[\protect\citename{Çelikkaya \bgroup et al.\egroup
  }2013]{celikkaya2013named}
Gökhan Çelikkaya, Dilara Torunoğlu, and Gülşen Eryiğit.
\newblock 2013.
\newblock Named entity recognition on real data: a preliminary investigation
  for turkish.
\newblock In {\em Application of Information and Communication Technologies
  (AICT), 2013 7th International Conference on}, pages 1--5. IEEE.

\bibitem[\protect\citename{Gerber \bgroup et al.\egroup }2013]{gerber2013real}
Daniel Gerber, Sebastian Hellmann, Lorenz B{\"u}hmann, Tommaso Soru, Ricardo
  Usbeck, and Axel-Cyrille~Ngonga Ngomo.
\newblock 2013.
\newblock Real-time rdf extraction from unstructured data streams.
\newblock In {\em The Semantic Web--ISWC 2013}, pages 135--150. Springer.

\bibitem[\protect\citename{Heck \bgroup et al.\egroup
  }2013]{heck2013leveraging}
Larry~P Heck, Dilek Hakkani-T{\"u}r, and G{\"o}khan T{\"u}r.
\newblock 2013.
\newblock Leveraging knowledge graphs for web-scale unsupervised semantic
  parsing.
\newblock In {\em INTERSPEECH}, pages 1594--1598.

\bibitem[\protect\citename{Hoffart \bgroup et al.\egroup
  }2011]{hoffart2011robust}
Johannes Hoffart, Mohamed~Amir Yosef, Ilaria Bordino, Hagen F{\"u}rstenau,
  Manfred Pinkal, Marc Spaniol, Bilyana Taneva, Stefan Thater, and Gerhard
  Weikum.
\newblock 2011.
\newblock Robust disambiguation of named entities in text.
\newblock In {\em Proceedings of the Conference on Empirical Methods in Natural
  Language Processing}, pages 782--792. Association for Computational
  Linguistics.

\bibitem[\protect\citename{K{\i}l{\i}n{\c{c}} \bgroup et al.\egroup
  }2015]{kilincc2015ttc}
Deniz K{\i}l{\i}n{\c{c}}, Ak{\i}n {\"O}z{\c{c}}ift, Fatma Bozyigit, Pelin
  Y{\i}ld{\i}r{\i}m, Fatih Y{\"u}calar, and Emin Borandag.
\newblock 2015.
\newblock Ttc-3600: A new benchmark dataset for turkish text categorization.
\newblock {\em Journal of Information Science}, page 0165551515620551.

\bibitem[\protect\citename{K{\"u}{\c{c}}{\"u}k and
  Steinberger}2014]{kuccuk2014experiments}
Dilek K{\"u}{\c{c}}{\"u}k and Ralf Steinberger.
\newblock 2014.
\newblock Experiments to improve named entity recognition on turkish tweets.
\newblock {\em arXiv preprint arXiv:1410.8668}.

\bibitem[\protect\citename{K{\"u}{\c{c}}{\"u}k \bgroup et al.\egroup
  }2014]{kuccuk2014named}
Dilek K{\"u}{\c{c}}{\"u}k, Guillaume Jacquet, and Ralf Steinberger.
\newblock 2014.
\newblock Named entity recognition on turkish tweets.
\newblock In {\em LREC}, pages 450--454.

\bibitem[\protect\citename{Lewis \bgroup et al.\egroup }2004]{lewis2004rcv1}
David~D Lewis, Yiming Yang, Tony~G Rose, and Fan Li.
\newblock 2004.
\newblock Rcv1: A new benchmark collection for text categorization research.
\newblock {\em The Journal of Machine Learning Research}, 5:361--397.

\bibitem[\protect\citename{Ling and Weld}2012]{ling2012fine}
Xiao Ling and Daniel~S Weld.
\newblock 2012.
\newblock Fine-grained entity recognition.
\newblock In {\em AAAI}.

\bibitem[\protect\citename{Liu \bgroup et al.\egroup }2015]{liu2015automated}
Qian Liu, Zhiqiang Gao, Bing Liu, and Yuanlin Zhang.
\newblock 2015.
\newblock Automated rule selection for aspect extraction in opinion mining.
\newblock In {\em International Joint Conference on Artificial Intelligence
  (IJCAI)}.

\bibitem[\protect\citename{Mendes \bgroup et al.\egroup
  }2011]{mendes2011dbpedia}
Pablo~N Mendes, Max Jakob, Andr{\'e}s Garc{\'\i}a-Silva, and Christian Bizer.
\newblock 2011.
\newblock Dbpedia spotlight: shedding light on the web of documents.
\newblock In {\em Proceedings of the 7th International Conference on Semantic
  Systems}, pages 1--8. ACM.

\bibitem[\protect\citename{Mukherjee and Liu}2010]{mukherjee2010improving}
Arjun Mukherjee and Bing Liu.
\newblock 2010.
\newblock Improving gender classification of blog authors.
\newblock In {\em Proceedings of the 2010 conference on Empirical Methods in
  natural Language Processing}, pages 207--217. Association for Computational
  Linguistics.

\bibitem[\protect\citename{Önal and Karagöz}2015]{onal2015named}
Kezban~Dilek Önal and Pinar Karagöz.
\newblock 2015.
\newblock Named entity recognition from scratch on social media.

\bibitem[\protect\citename{Pasca \bgroup et al.\egroup
  }2006]{pasca2006organizing}
Marius Pasca, Dekang Lin, Jeffrey Bigham, Andrei Lifchits, and Alpa Jain.
\newblock 2006.
\newblock Organizing and searching the world wide web of facts-step one: the
  one-million fact extraction challenge.
\newblock In {\em AAAI}, volume~6, pages 1400--1405.

\bibitem[\protect\citename{Rose \bgroup et al.\egroup }2002]{rose2002reuters}
Tony Rose, Mark Stevenson, and Miles Whitehead.
\newblock 2002.
\newblock The reuters corpus volume 1-from yesterday's news to tomorrow's
  language resources.
\newblock In {\em LREC}, volume~2, pages 827--832.

\bibitem[\protect\citename{Sebastiani}2002]{sebastiani2002machine}
Fabrizio Sebastiani.
\newblock 2002.
\newblock Machine learning in automated text categorization.
\newblock {\em ACM computing surveys (CSUR)}, 34(1):1--47.

\bibitem[\protect\citename{Singh \bgroup et al.\egroup }2011]{singh2011large}
Sameer Singh, Amarnag Subramanya, Fernando Pereira, and Andrew McCallum.
\newblock 2011.
\newblock Large-scale cross-document coreference using distributed inference
  and hierarchical models.
\newblock In {\em Proceedings of the 49th Annual Meeting of the Association for
  Computational Linguistics: Human Language Technologies-Volume 1}, pages
  793--803. Association for Computational Linguistics.

\bibitem[\protect\citename{Steinmetz \bgroup et al.\egroup
  }2013]{steinmetz2013statistical}
Nadine Steinmetz, Magnus Knuth, and Harald Sack.
\newblock 2013.
\newblock Statistical analyses of named entity disambiguation benchmarks.
\newblock In {\em NLP-DBPEDIA@ ISWC}.

\bibitem[\protect\citename{Tantug}2015]{tantugrecognizing}
Beyza Ekenand A~C{\"u}neyd Tantug.
\newblock 2015.
\newblock Recognizing named entities in turkish tweets.

\bibitem[\protect\citename{Tatar and Çiçekli}2011]{tatar2011automatic}
Serhan Tatar and İlyas Çiçekli.
\newblock 2011.
\newblock Automatic rule learning exploiting morphological features for named
  entity recognition in turkish.
\newblock {\em Journal of Information Science}, 37(2):137--151.

\bibitem[\protect\citename{Tjong Kim~Sang and
  De~Meulder}2003]{tjong2003introduction}
Erik~F Tjong Kim~Sang and Fien De~Meulder.
\newblock 2003.
\newblock Introduction to the conll-2003 shared task: Language-independent
  named entity recognition.
\newblock In {\em Proceedings of the seventh conference on Natural language
  learning at HLT-NAACL 2003-Volume 4}, pages 142--147. Association for
  Computational Linguistics.

\bibitem[\protect\citename{Toraman \bgroup et al.\egroup
  }2011]{toraman2011developing}
Cagri Toraman, Fazl{\i} Can, and Seyit Ko{\c{c}}berber.
\newblock 2011.
\newblock Developing a text categorization template for turkish news portals.
\newblock In {\em Innovations in Intelligent Systems and Applications (INISTA),
  2011 International Symposium on}, pages 379--383. IEEE.

\bibitem[\protect\citename{T{\"u}r \bgroup et al.\egroup
  }2003]{tur2003statistical}
G{\"o}khan T{\"u}r, Dilek Hakkani-T{\"u}r, and Kemal Oflazer.
\newblock 2003.
\newblock A statistical information extraction system for turkish.
\newblock {\em Natural Language Engineering}, 9(02):181--210.

\bibitem[\protect\citename{Yogatama \bgroup et al.\egroup
  }2015]{yogatama2015embedding}
Dani Yogatama, Dan Gillick, and Nevena Lazic.
\newblock 2015.
\newblock Embedding methods for fine grained entity type classification.
\newblock In {\em Proceedings of the 53rd Annual Meeting of the Association for
  Computational Linguistics and the 7th International Joint Conference on
  Natural Language Processing of the Asian Federation of Natural Language
  Processing, ACL}, pages 26--31.

\bibitem[\protect\citename{Yosef \bgroup et al.\egroup }2011]{yosef2011aida}
Mohamed~Amir Yosef, Johannes Hoffart, Ilaria Bordino, Marc Spaniol, and Gerhard
  Weikum.
\newblock 2011.
\newblock Aida: An online tool for accurate disambiguation of named entities in
  text and tables.
\newblock {\em Proceedings of the VLDB Endowment}, 4(12):1450--1453.

\end{thebibliography}
}	
	
\end{document}